\documentclass[fleqn,10pt,twocolumn]{article}
\usepackage{graphicx}
\usepackage[inline,shortlabels]{enumitem}
\usepackage{cite}
\usepackage{amsmath,amssymb,amsfonts}
\usepackage{algorithm}
\usepackage[noend]{algpseudocode}
\usepackage{textcomp}
\usepackage{xcolor}
\usepackage{color,soul}
\usepackage{indentfirst}
\usepackage{svg}
\usepackage{cite}
\usepackage{amsmath,amssymb,amsfonts}
\usepackage{graphicx}
\usepackage{textcomp}
\usepackage{xcolor}

\usepackage{bm}
\usepackage{dsfont}
\usepackage{amsmath}
\usepackage{subfig}
\usepackage{flushend}

\def\BibTeX{{\rm B\kern-.05em{\sc i\kern-.025em b}\kern-.08em
		T\kern-.1667em\lower.7ex\hbox{E}\kern-.125emX}}
\title{Obstacle-Aware Navigation of Soft Growing Robots via Deep Reinforcement Learning}
	\author{{Haitham El-Hussieny}$^{1}$, {Ibrahim Hameed}$^{2}$\\
	$^1$ Department of Mechatronics and Robotics Engineering,\\ Egypt-Japan University of Science and Technology, Alexandria, Egypt\\
		$^2$ Department of ICT and Natural Sciences,\\ Norwegian University of Science and Technology, 7034 Trondheim,  Norway\\
haitham.elhussieny@ejust.edu.eg, ibib@ntnu.no
 }

\begin{document}
	\graphicspath{{Pictures/}}

\maketitle
\begin{abstract}
	Soft growing robots, are a type of robots that are designed to move and adapt to their environment in a similar way to how plants grow and move with potential applications where they could be used to navigate through tight spaces, dangerous terrain, and hard-to-reach areas. This research explores the application of deep reinforcement Q-learning algorithm for facilitating the navigation of the soft growing robots in cluttered environments. The proposed algorithm utilizes the flexibility of the soft robot to adapt and incorporate the interaction between the robot and the environment into the decision-making process. Results from simulations show that the proposed algorithm improves the soft robot's ability to navigate effectively and efficiently in confined spaces. This study presents a promising approach to addressing the challenges faced by growing robots in particular and soft robots general in planning obstacle-aware paths in real-world scenarios.
\end{abstract}

\section{Introduction}
The exploration of confined spaces, such as those encountered in Minimally Invasive Surgeries (MIS) or the inspection of archaeological sites, presents significant challenges for traditional rigid robot designs \cite{leibrandt2017concentric, brantnercontrolling}. Consequently, there is a critical need for innovative materials and locomotion systems in robotics to navigate these challenging environments effectively. Drawing inspiration from biological systems like elephant trunks, octopus tentacles, and snakes, the development of soft continuum robots featuring continuous bending backbones has facilitated non-destructive navigation in congested spaces \cite{webster2010design, greer2018obstacle, seleem2019guided}. However, the limited lengths of traditional continuum robots restrict their capacity to explore more distant spaces \cite{liljeback2012review}.

Addressing this limitation, the concept of growth mobility, inspired by the way plants grow, has emerged as a groundbreaking approach in robotics. Growing robots, emulating the biological growth of plants, can extend their lengths, volumes, or knowledge, gradually adapting to their environment \cite{del2018toward}. These robots, made of soft materials or equipped with flexible joints, can extend to reach faraway spaces while maintaining compliance in confined settings. Prior efforts in designing long, flexible robots suitable for tight spaces include the “Active Hose” by Tsukagoshi et al., a multi-degree-of-freedom robot designed for search and rescue applications \cite{tsukagoshi2001active}. Isaki et al. developed a flexible, extended cable with a camera for exploration in narrow areas \cite{isaki2006development}, and the “Slime Scope,” a pneumatically driven soft arm, was created for use in rubble environments \cite{mishima2003development}. However, these designs often require moving the entire robot body, leading to significant friction with the environment.

Recent advancements in plant-inspired robotics have led to the development of two novel types of growing robots capable of extending their body lengths by adding materials at their tips. Sadeghi et al. proposed a plant root-like robot with a 3D printer-like head at its tip, enabling efficient steering by varying material deposition speeds \cite{sadeghi2017plant, sadeghi2020passive}. However, the growth speed is environment-dependent. In contrast, Hawkes et al. developed a vine-like growing robot using a "tip-eversion" mechanism \cite{Hawkes2017, blumenschein2017modeling}. This pneumatically-driven robot, made from thin-walled polyethylene tubing, can extend for tens of meters and navigate slippery or sticky environments. Steering is achieved through environmental interaction or by using Series Pneumatic Artificial Muscles (SPAM) for controlled movements \cite{greer2018obstacle, el2018development}. The unique combination of lengthening capacity, length-to-diameter ratio, and a compliant body allows vine-like robots to effectively navigate through distant, cluttered environments, as demonstrated in \cite{coad2019vine, haitham2019dynamic}.

The very features that make vine-like robots so versatile also introduce complex challenges in motion planning. One of the critical aspects of vine robots is their irreversible growth process. Once a segment of the robot has extended or turned in a particular direction, retracting or reversing this action is not feasible. This irreversible nature of growth necessitates highly accurate and forward-thinking motion planning. In their recent work, \cite{el2023plant} have pioneered the use of a Model Predictive Control (MPC) approach in the context of vine robot navigation, incorporating the kinematics of a vine robot as the predictive model, enabling advanced motion planning and obstacle avoidance. However, the development of an accurate kinematic model for vine-growing robots poses significant challenges.

\begin{figure}[!t!p]
	\centering
	\includegraphics[width=\columnwidth]{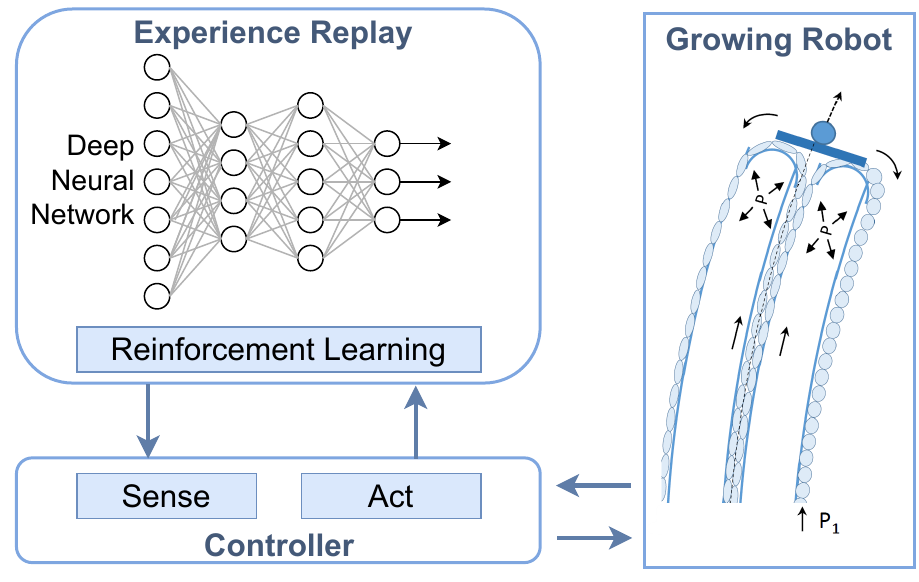}
	\caption{Enhancement of Movement Skills in the Vine-Growing Robot Through Deep Reinforcement Learning: This process involves the robot's adaptive learning from its interactions with the environment.}
	\label{fig:learn}	
\end{figure}

This research introduces a new application of deep reinforcement learning, specifically Deep Q Networks (DQN), to enhance the navigational capabilities of soft growing robots, as depicted in Figure \ref{fig:learn}. Our algorithm capitalizes on the inherent flexibility of soft robots, seamlessly integrating environmental interactions into the decision-making process. This approach is particularly geared towards enabling navigation through densely cluttered spaces, a task that presents significant challenges for conventional robotic systems.

The findings from our simulations indicate a marked improvement in the ability of soft robots to exploit obstacles effectively to reach challenging goals, leveraging the DQN framework. This advancement addresses a critical obstacle in soft robotics, significantly enhancing the operational performance of soft growing robots in real-life scenarios, especially those characterized by complex, obstacle-rich environments. By equipping these robots with the capability to plan and execute obstacle-aware paths more effectively, this research lays the groundwork for their enhanced utilization in diverse, real-world applications.

The paper is organized as follows: Section \ref{sec:model} introduces the kinematic model of the vine growing robot, including the interaction model with obstacles. In Section \ref{sec:dqn} details the proposed DQN reinforcement learning algorithm, including the observation, action and reward definitions. In Section \ref{sec:results}, we present the simulation results to asses the performance of our DQN approach for planning safe paths for growing robots. Finally, Section \ref{sec:conc} concludes the paper, summarizes our main contributions, and outlines potential directions for future work.

\section{Modeling of Vine Robots} 
\label{sec:model}
\begin{figure}[!t!p]
	\centering
	\includegraphics[width=\columnwidth, trim={5cm 4cm 9cm 1cm},clip]{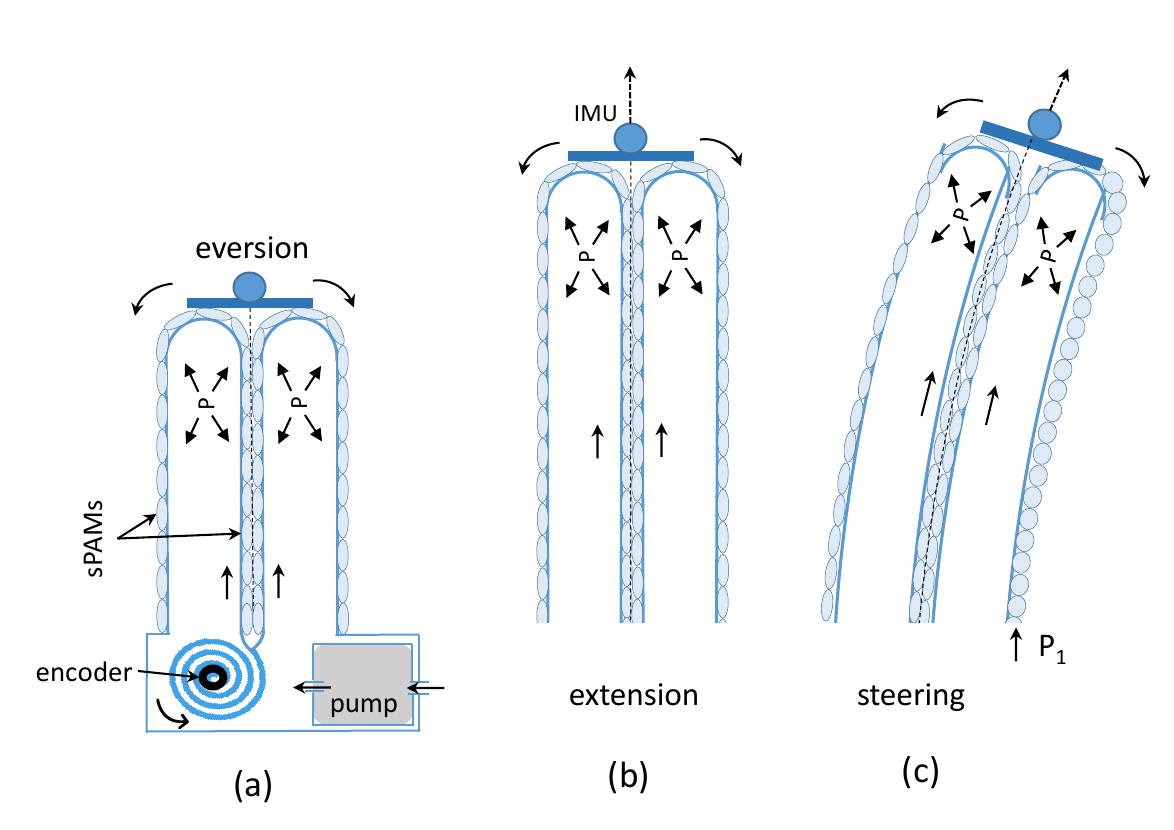}
\caption{Operational Mechanics of the Growing Vine Robot. (a) The application of air pressure to the robot's central tube aids in extending the tip, as shown in (b). The steering mechanism, depicted in (c), is achieved by altering the air pressure in one or more of the soft pneumatic artificial muscles (sPAMs) surrounding the vine robot.}
	\label{fig:robot}	
\end{figure}

In this research, we focus on the vine-growing robot recently developed by \cite{coad2019vine}. Using an eversion mechanism \cite{greer2018soft}, the robot can elongate its tip up to tens of meters. The robot's body is made of a thin-walled polyethylene tube, initially inverted, as depicted in Figure \ref{fig:robot}. The robot's length can be increased by applying air pressure to its chamber, which permits the tip to move away from its base. The eversion mechanism allows the vine robot to navigate easily through adhesive environments without getting stuck. To control the robot's bending, three serial Pneumatic Actuator Muscles (sPAM) are installed around its circumference, and air pressure is applied to these actuators. The arrangement of these sPAM actuators enables 360$^\circ$ steering, achieved by manipulating the input pressure for each actuator, as described in \cite{coad2019vine}.

This section is dedicated to exploring the connection between actuation parameters and the position of the robot's tip, as delineated by the kinematic model. Additionally, it will illustrate the method of integrating the robot body's interaction with obstacles into the kinematics. This integration will be achieved by resolving a nonlinear constrained optimization problem that symbolizes this interaction.

\subsection{Kinematics of Vine Robots}
Given the inherent characteristics of soft growing robots, particularly their lightweight construction and the typically slow-paced nature of their operational movements, the primary focus in research and application tends to be on statics and kinematics, rather than dynamics. The forward kinematics of the vine robot are formulated predominantly under the guiding principle of the constant-curvature assumption, as detailed in the seminal work of Jones et al. (2006) \cite{jones2006kinematics}. This robot is conceptualized as a single-section, extensible entity exhibiting continuum-like characteristics, endowed with dual degrees of freedom for curvature and bending, along with an additional degree for axial extension.

\begin{figure}[!t!p]
	\centering
	\includegraphics[width=.8\columnwidth,keepaspectratio]{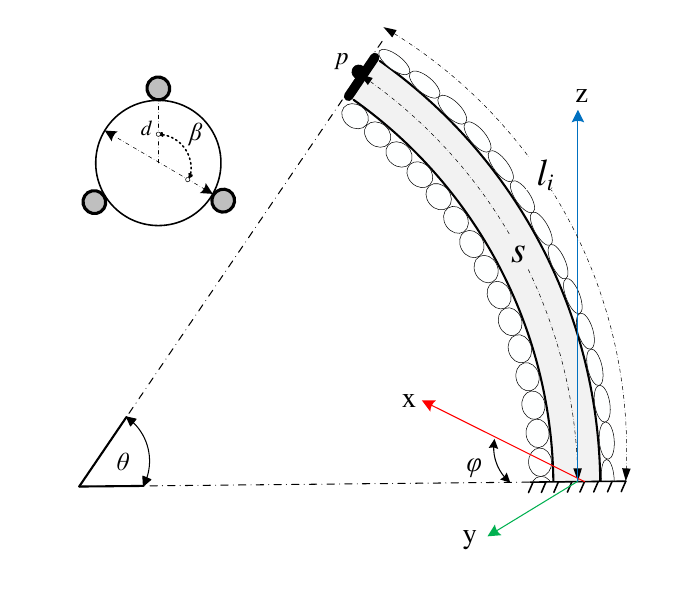}
	\caption{Schematic of vine-like growing robot and its configuration parameters. The robot is characterized by its length $s$, curvature $\kappa = \frac{\theta}{s}$, and angle of robot plane $\phi$.}
	\label{fig:robot_kin}	
\end{figure}

The spatial orientation and positioning of the robot, denoted as $\mathbf{T}_r^b$, are intrinsically linked to its configurational state, represented by $\bm{q} \in \mathds{R}^3$. This state vector $\bm{q}$ encompasses the pivotal elements defining the robot's structure: the length of the robot ($s$), the curvature ($\kappa$), and the angle of the curvature plane ($\phi$), all of which are comprehensively illustrated in Figure \ref{fig:robot_kin}. The pose $\mathbf{T}_r^b$ is derived from these parameters, offering a precise mathematical representation of the robot’s positioning and orientation in space.

{\small
\begin{equation}
	\begin{aligned}
			\label{eq:kinem}
		& \mathbf{A}=\left[\begin{array}{cc}
			\cos ^2 \phi(\cos \kappa s-1)+1 & \sin \phi \cos \phi(\cos \kappa s-1) \\
			\sin \phi \cos \phi(\cos \kappa s-1) & \cos ^2 \phi(1-\cos \kappa s)+\cos \kappa s \\
			\cos \phi \sin \kappa s & \sin \phi \sin \kappa s \\
			0 & 0
		\end{array}\right. \\
		& \left.\begin{array}{cc}
			-\cos \phi \sin \kappa s & \frac{\cos \phi(\cos \kappa s-1)}{\kappa} \\
			-\sin \phi \sin \kappa s & \frac{\sin \phi(\cos \kappa s-1)}{\kappa} \\
			\cos \kappa s & \frac{\sin \kappa s}{\kappa} \\
			0 & 1
		\end{array}\right] . \\
		&
	\end{aligned}
\end{equation}
}
The robot's tip position $\bm{p} = [x, y, z]^T \in \mathds{R}$$^3$ in Cartesian space can be stated from Eq. \eqref{eq:kinem} as,

\begin{equation}
	\label{eq:taylor}
	\begin{aligned}
		x & = \frac{\cos \phi(\cos \kappa s-1)}{\kappa}, \\
		y & = 	\frac{\sin \phi(\cos \kappa s-1)}{\kappa}, \\
		z & = \frac{\sin \kappa s}{\kappa}
	\end{aligned}
\end{equation}

In the context of this research, we have adopted the simplifying assumption of a planar environment, denoted mathematically as $\phi = 0$. This assumption is primarily for the sake of analytical and computational simplicity, allowing us to focus on the core aspects of the algorithm and control strategy without the additional complexity introduced by a three-dimensional environment. However, it is important to note that the extension of our model and methodologies to a spatial, three-dimensional environment is straightforward. 

The kinematic formulation encapsulated in Eq. \eqref{eq:taylor} is key for understanding the vine robot's intrinsic behavior and practical application in autonomous navigation. This model is integral to the training regimen of the Deep Q-Network (DQN) agent, a critical component in enabling the vine robot to accurately navigate towards designated target positions, particularly in scenarios involving interaction with its surrounding environment.

Central to the operational mechanics of the vine robot is the actuation lengths, denoted by $\bm{l} = [s, l_1, l_2, l_3]$. These lengths represent the tangible actuation space, encompassing both the core robot length and the lengths of the soft Pneumatic Artificial Muscles (sPAMs). Notably, utilizing shape parameters within this framework does more than merely dictate the robot's movements; it effectively generalizes the control problem. This generalization ensures that the control strategies developed are not limited to this specific vine robot but apply to a broader spectrum of continuum-like robots adhering to the constant-curvature model previously discussed.

\subsection{Interaction Modeling}
\begin{figure}[!b!p]
	\centering
	\includegraphics[clip, trim=0.8cm 5.2cm 4.4cm 0cm, width = \columnwidth]{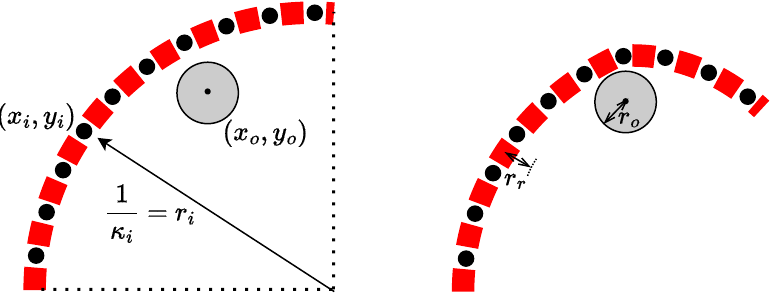}
	\caption{Effect of a point obstacle on the continuum robot shape. On the left the robot has no interaction with the obstacle and the robot curvature is the same to all segments. On the right, the robot has collided with a point obstacle making the robot curvature not consistent along it's backbone.}
	\label{fig:robot_obstacle}
\end{figure}

When a vine robot makes contact with an obstacle, it adapts its form to accommodate this interaction. Predicting and understanding this shape adaptation is essential for refining control algorithms and ensuring smooth, efficient navigation.

In order to integrate the influence of obstacles into the change of the robot's shape, we implemented the strain energy minimization principle as emphasized in \cite{ashwin2021profile}. In general, the approach involves treating a single-section continuum robot of length $s$ as a flexible beam formed of $N$ sub-segments. Each segment is subjected independently to bending deformation caused by the external torque resulting the actuation and the interaction with the environment as depicted in Figure \ref{fig:robot_obstacle}. Consequently, the robot assumes a configuration that optimizes its strain energy at a state of static equilibrium. This optimization leads to a curvature alignment denoted as $\bm{\kappa} \in \mathds{R}^N$ across the segmented length of the vine robot's body, divided into $N$ sections. This alignment serves as the central objective, encapsulated by the following key objective function with the aim of finding the set of curvatures $\bm{\kappa}\in \mathds{R}^N$ for the sub-segments.

\begin{equation}
	\label{eq:optimization}
	\arg \min_{\bm{\kappa}} \left( \gamma_1\sum_{i=0}^{N-1} (\kappa_{i} - \kappa_t)^2 + \gamma_2 \sum_{i=0}^{N-2} (\kappa_{i+1} - \kappa_i)^2 \right) 
\end{equation}
trim={5cm 4cm 9cm 1cm},clip
In the specified equation, the formulation is strategically divided into two principal terms, each addressing a distinct aspect of the vine robot's curvature control. The first term is designed to minimize the deviation between the curvature of each sub-segment of the robot and a predefined target curvature, $\kappa$. This target curvature is established under the assumption of obstacle-free conditions, serving as a benchmark for the desired trajectory of the robot in an unobstructed environment.

Simultaneously, the second term in the equation is tasked with a different objective: it aims to reduce the disparity in curvature between adjacent segments of the robot. This is a critical consideration, as it adheres to the principle of minimizing strain energy, thereby preserving the integrity of the robot's continuous backbone structure. The necessity to maintain a smooth transition between segments is paramount in ensuring the robot's efficient and effective movement, particularly in scenarios involving complex navigation paths.

The incorporation of the factors $\gamma_1$ and $\gamma_2$ in this equation plays a vital role. These weighting factors are employed to establish a balance between the dual objectives of the equation.

To avoid the growing robot from colliding with the obstacle, the optimization problem in Eq. \eqref{eq:optimization} is subjected to the constraint that the Cartesian coordinates $(x_i, y_i)$ of the tip of each sub segment $i$ is within a safe distance $r_r$ from the center of the obstacle at the location $(x_o, y_o)$ with radius $r_o$ as follow, assuming the obstacle location is known

\begin{equation}
\label{eqLconstraint}
d_{io} \geq r_o+ r_r, \quad \forall i = 0, 1, \ldots, N-1
\end{equation}
\noindent where $d_{io} = \sqrt{(x_i - x_o)^2+(y_i - y_o)^2}$ is the Euclidean distance from the segment $i$ of the robot and the obstacle location. 

While the constant curvature model serves as the foundational assumption in our kinematic modeling of the vine robot, it's important to recognize that interactions with environmental obstacles can significantly alter the robot's shape, thereby challenging this assumption. In scenarios where the robot comes into contact with an obstacle, its behavior diverges from that of a single-segment, constant curvature entity. Instead, the robot begins to exhibit characteristics akin to a multi-segmented structure, displaying varying curvatures along its length.

\section{DQN Reinforcement Learning Algorithm}
\label{sec:dqn}
This section elaborates on the Reinforcement Learning (RL) algorithm employed to train the growing robot. The primary objective is to enable the robot to efficiently navigate towards a specified goal within its environment, while skillfully navigating around and interacting with obstacles. The RL agent operates by observing the current state $\bm{x} \in \mathds{R}^9$ of the growing robot, which is defined within the observation space as:

\begin{equation}
	\bm{x} = [s\ \kappa\ \dot{s}\ \dot{\kappa}\ x_g\ y_g\ x_o\ y_o\ d]^T
\end{equation}

In this state vector $\bm{x}$, $s$ and $\kappa$ represent the robot's length and curvature, respectively, and $\dot{s}$ and $\dot{\kappa}$ are the corresponding time derivatives of these quantities. These variables are crucial for the RL agent to understand and predict the robot's physical configuration and movement dynamics. To maintain awareness of the goal and obstacle locations within the environment, the coordinates $(x_g, y_g)$ for the goal and $(x_o, y_o)$ for the obstacle are included in the observation space. Additionally, the Euclidean distance $d$ is calculated and tracked. This distance, defined between the robot's tip $(x, y)$ as derived in Eq. \eqref{eq:taylor} and the goal position, is given by:
\begin{equation}
	\label{eq:error}
	d = \sqrt{(x - x_g)^2 + (y - y_g)^2}
\end{equation}

In order to ensure the generalizability of the RL model to different goal locations, the position of the goal is varied randomly during the training process, with a probability of $0.2$ for changing in each training iteration. This randomization strategy is designed to expose the growing robot to a diverse set of scenarios, thereby enhancing its ability to adapt and perform effectively in a wide range of environmental conditions and goal configurations.

The reinforcement learning agent in this study is designed to manage the movement of a growing robot by sampling an action vector $\bm{a}$, defined as follows:

\begin{equation}
	\bm{a} = [\dot{s}\ \dot{\kappa}]
\end{equation}

Here, $\dot{s}$ represents the robot's growth speed, and $\dot{\kappa}$ indicates the rate of change in the robot's curvature. By incorporating these specific actuation variables, the proposed algorithm demonstrates versatility and adaptability across a broad range of continuum-like robotic systems. This flexibility is crucial as it ensures the applicability of the goal-reaching algorithms irrespective of the underlying actuation mechanism, whether it be cable-driven or pneumatically driven systems.

In the domain of Q-learning, a crucial aspect is the representation of actuation in discrete values. This discretization ensures that each actuation component of the robot operates within a predefined set of actions, effectively covering the entire operational range. Specifically, in our model:

\begin{enumerate}
	\item The rate of change of curvature, $\dot{\kappa}$, is restricted to three discrete values, $[\kappa_l\ 0\ \kappa_r]$ rad/s. These values correspond to the robot bending to the left, remaining stationary (no bending), and bending to the right, respectively.
	
	\item The growth speed, $\dot{s}$, is either set to elongate the robot at a constant rate of $s_{m}$ m/s or to halt its elongation, denoted as $\dot{s} = 0$. 
\end{enumerate}

Consequently, the action space for the robot is defined as a set of all possible combinations of these discrete values:

$$\bm{a} = \left\{\begin{bmatrix} \kappa_l\\0\end{bmatrix}, \begin{bmatrix} \kappa_l\\s_{m}\end{bmatrix},\begin{bmatrix} 0\\0\end{bmatrix},\begin{bmatrix} 0\\s_{m}\end{bmatrix},\begin{bmatrix}\kappa_r\\0\end{bmatrix},\begin{bmatrix}\kappa_r\\s_{m}\end{bmatrix}\right\} $$

This discrete action space encapsulates all the possible movements of the robot within its operational constraints, allowing for a structured approach in the Q-learning algorithm to optimize the robot's behavior for achieving specific goals.
At the onset of each training episode, the RL algorithm initializes with predetermined parameters $\kappa_0$ and $s_0$. The algorithm initially adopts a strategy of exploration, where actions $\bm{a}$ are chosen randomly based on an exploration probability $\epsilon$, as delineated in Algorithm \ref{alg:action}. This exploratory phase is crucial for the agent to acquire diverse experiences and insights into the environment's dynamics.

As training progresses, the algorithm gradually shifts its focus towards exploitation, increasingly selecting actions that are anticipated to yield the highest cumulative reward. The formulation for calculating the cumulative reward $R_k$ at each step $k$ is expressed as:

\begin{equation}
	\label{eq:acc_reward}
	R_k = r_k + \gamma r_{k+1} + \gamma^2 r_{k+2} + \ldots + \gamma^{n-k} r_n = r_k + R_{k+1}
\end{equation}

In this equation, $r_k$ represents the immediate reward assigned at each step $k$, and $\gamma$ is the discount factor, which balances the importance of immediate and future rewards.

The computation of the immediate reward $r_k$ in our proposed method is designed to incentivize goal-oriented behavior. Specifically, a high reward of +1000 is assigned if the robot's tip is within a maximum distance of $d_{\text{max}} = 0.1$ from the goal, leading to the termination of the training episode. Conversely, if the robot's tip is farther from the goal, the reward is inversely proportional to this distance $d$, as defined in Eq. \eqref{eq:error}. The reward function is thus formulated as:

\begin{equation}
	r_k = 1 - \left( \dfrac{\ln(1 + d_k)}{\ln(1 + d_{\text{max}})} \right)
\end{equation}

This reward structure ensures that the agent is motivated to minimize the distance to the goal, effectively guiding its learning process towards efficient and goal-directed actions. 

During both the training and testing phases of our approach, the position of the obstacle is assumed to be fixed. This static positioning of the obstacle simplifies the learning and evaluation process, allowing the reinforcement learning agent to focus on understanding and adapting to a consistent environmental factor. However, it's noteworthy that our approach possesses the inherent capability to handle dynamic obstacles as well. This adaptability stems from the inclusion of the goal position in the agent's observation space. Since the agent is already equipped to process and respond to changes in the goal's location, extending this adaptability to accommodate moving obstacles is a feasible and logical progression.

\begin{figure}[!t!p]
	\centering
	\includegraphics[width=.8\columnwidth,keepaspectratio]{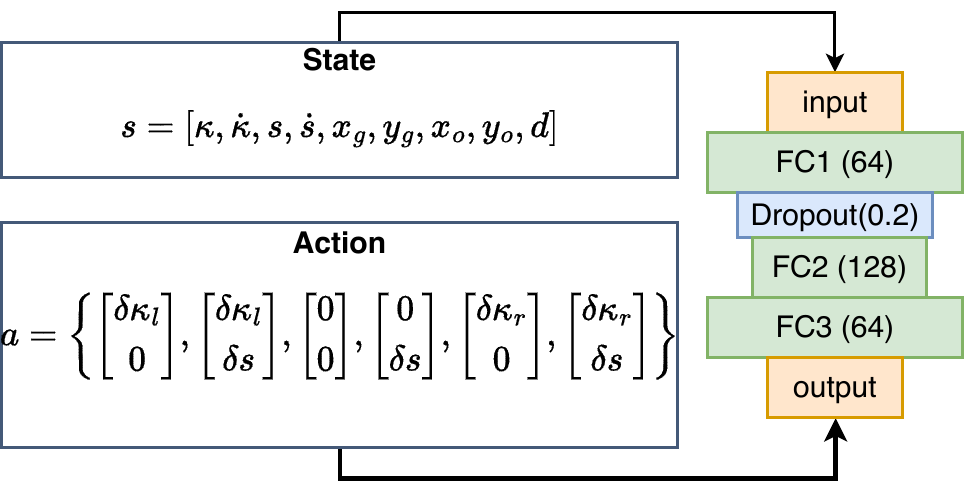}
	\caption{Illustration of the DQN Agent Architecture: This diagram depicts the agent as it observes state $s$ and executes action $a$. The process of transforming the observed state into an actionable decision is facilitated through a feed-forward neural network, which serves as the core mechanism for mapping observations to corresponding actions.}
	\label{fig:dqn}	
\end{figure}

The deep Q-learning algorithm \cite{mnih2015human} is utilized in this research to train the agent to reach the desired goal while maximizing the accumulative reward of Eq. \eqref{eq:acc_reward}. The RL agent learns to select the best possible action $a$ at every state $s$ as described in Algorithm \ref{alg:learn} and Figure \ref{fig:dqn}. This is represented by a policy $\pi(s)$
\begin{equation}
\pi(s) = \max_{a} Q(s,a)
\end{equation}

where $Q(s, a)$ is approximated in Q-learning algorithm as

\begin{equation}
\label{eq:q_value}
Q(s,a) \leftarrow  Q(s,a) + \alpha\left( r + \gamma \max_{a^{\prime}} Q(s^{\prime},a^{\prime}) - Q(s, a)  \right)
\end{equation}

where $s^\prime$ is the next state while $a^\prime$ is the next action. Meanwhile, $\alpha$ is the learning rate.

\begin{algorithm}[!t!p]
	\caption{Action sampling}\label{alg:action}
	\begin{algorithmic}
		\State $\bm{P} \gets \bm{P}_0$ 
		\For{episode = 1 to $N$}
		\For{$k$ = 1 to $M$} 
		\State with probability $\epsilon$ select random action $\bm{a}_k$, otherwise $\bm{a}_k = \max_{a} Q(s, a)$
		\State ($\bm{s}_{k+1}, r_k$) = \textbf{step}($\bm{s}_{k}, \bm{a}_{k}$) \Comment{Simulate the system}
		\State $\bm{P}_k = (\bm{s}_{k}, \bm{a}_{k}, \bm{s}_{k+1}, r_{k})$
		\EndFor
		\EndFor
	\end{algorithmic}
\end{algorithm}

\begin{algorithm}[!t!p]
	\caption{Learning Q-value}\label{alg:learn}
	\begin{algorithmic}
		\State $\bm{\theta} \gets \bm{\theta}_0$ \Comment{Initialize network weights}
		\While{No termination signal recieved}
		
		\State Sample random minibatch of ($\bm{s},\bm{a}, r, \bm{s^\prime}$) from $\bm{P}$.
		\State Set target = $r + \gamma \max_{a^\prime} Q_{\theta}(s^\prime, a^\prime)$
		\State Set prediction = $Q_{\theta}(s^\prime, a^\prime)$
		\State 	$L = \frac{1}{2} \left[\text{target} - \text{prediction}\right]^2$
		\State $\theta \leftarrow \theta - \alpha \dfrac{\partial L}{\partial \theta}$ \Comment{Update network weights}
		\EndWhile
		
	\end{algorithmic}
\end{algorithm}

In conventional Q-learning \cite{mnih2013playing}, where the state space is small, the Q value is represented as a table of state and the corresponding Q of each action in that state. However, in this navigation problem, the state space is enormous. Thus, a deep neural network is used to approximate the the Q-table as $Q_\theta(s, a)$, where $\theta$ is the deep neural network that is trained as follows:
\begin{equation}
\theta \leftarrow\theta - \alpha \dfrac{\partial L}{\partial \theta} 
\end{equation}

where $L$ is the loss:
\begin{equation}
	\begin{aligned}
	L &= \frac{1}{2} \left[\text{target} - \text{prediction}\right]^2\\
	& = \frac{1}{2} \left[ r + \gamma \max_{a^{\prime}} Q(s^{\prime},a^{\prime}) - Q(s, a) \right]^2
	\end{aligned}
\end{equation}

\section{Results and Discussion}
\label{sec:results}
\begin{figure}[!t!p]
	\centering
	\includegraphics[width=\columnwidth]{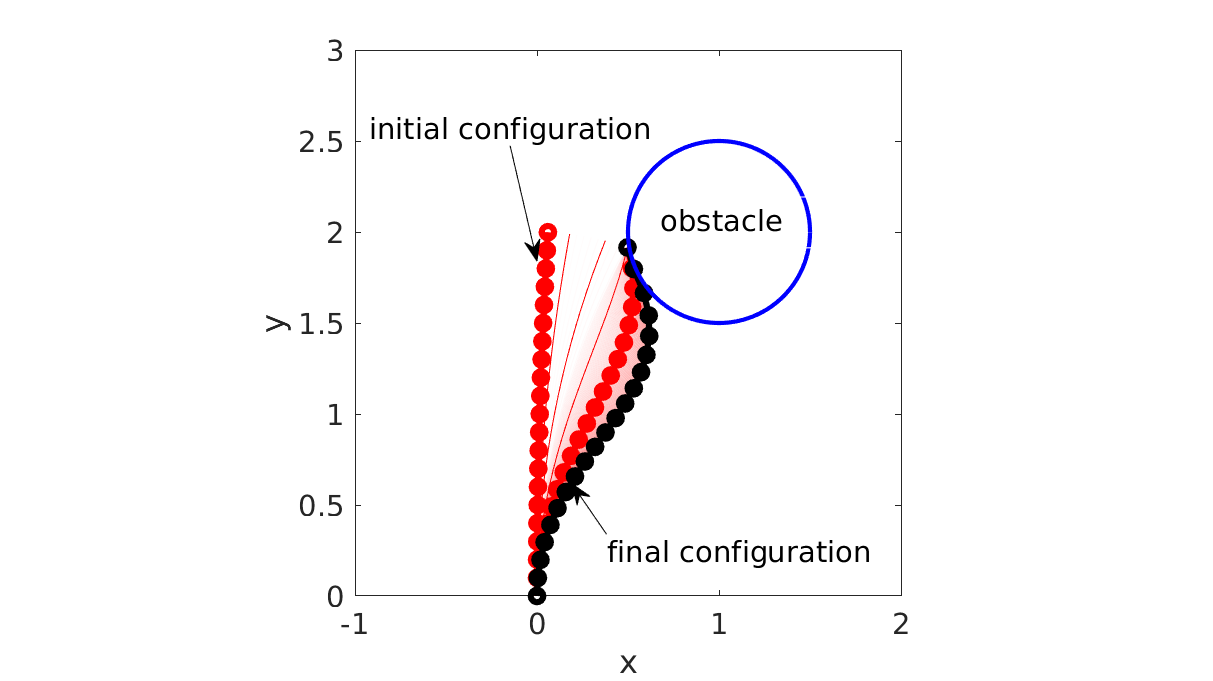}
	\caption{The vine growing robot interacting with an obstacle after applying constant actuation of $\dot{\kappa}$.}
	\label{fig:obstacle}		
\end{figure}

A comprehensive series of numerical experiments designed to evaluate the effectiveness of the DQN-based method for obstacle-aware navigation in growing robots. Initially, the focus was on analyzing the interaction model. This model crucially predicts how the morphology of the vine robot alters in response to encounters with obstacles. In Figure \ref{fig:obstacle}, we illustrate the initial phase of the robot's navigation sequence, wherein it begins in a configuration that is free from any collision with the obstacle, depicted in transparent red. Subsequently, as the robot navigates through its environment, there is an incremental increase in its bending motion, applying constant $\dot{\kappa}$. This progression continues until the point of collision with the obstacle. Upon this interaction, the robot's shape undergoes a transformation, conforming to the parameters set out by the interaction model discussed in Section \ref{sec:model}. Such an analysis is vital for understanding the adaptive behavior of the robot under varying environmental constraints.

In this study, we configured the replay memory to accommodate a maximum of 20,000 steps, as delineated in Algorithm \ref{alg:action}. This memory utilizes a First-In-First-Out (FIFO) strategy for data replacement, ensuring a continual update of experiences for the learning agent. The training protocol, detailed in Algorithm \ref{alg:learn}, comprised an extensive series of 200,000 episodes ($N$), with each episode constrained to a maximum of 100 steps ($k_{\text{max}}$). This comprehensive approach is designed to provide the agent with a wide array of scenarios for robust training.

The learning thread, initiating after a prelude of 20 episodes, facilitates the accumulation of initial experiences in the replay memory, essential for effective learning. A batch size of 64 was chosen for the random selection of experiences from the memory pool (denoted as P), balancing the computational efficiency and diversity in experience sampling.

The exploration rate of the network was initially set to 1.0, promoting a purely exploratory approach at the onset of the learning phase. Subsequently, this rate was decremented by 0.05 following each episode, culminating at zero. Such a gradual decrement facilitates a strategic shift from exploration to exploitation, enabling the agent to progressively rely on learned behaviors over random actions.

For optimization, the ADAMS optimizer \cite{zeiler2012adadelta} was employed, renowned for its efficacy in handling large datasets and complex variable interactions. Finally, the discount factor and learning rate were set at 0.5 and 0.1, respectively. 

\subsection{Fixed goal, obstacle-free}
\begin{figure}[!t!p]
	\fontsize{40}{18}\selectfont   
		\centering
	\includegraphics[width=\columnwidth]{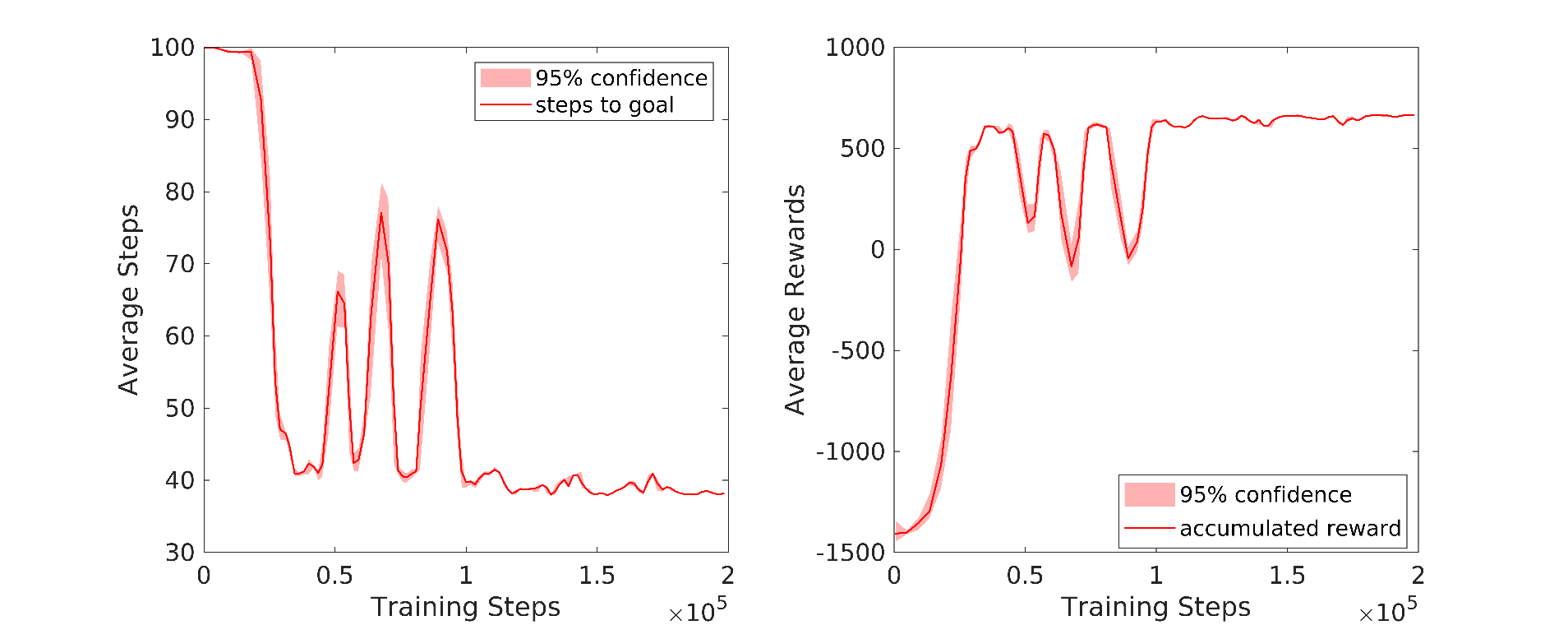}
	\caption{Evaluation of the Proposed DQN Goal-Seeking Algorithm: This figure illustrates the learning performance of the algorithm, showcasing two key metrics. On the left, the graph displays the number of steps required by the agent to reach the goal. On the right, the graph depicts the accumulated reward achieved by the agent.}
	\label{fig:reward}	
	\normalsize                    
\end{figure}
In the initial phase, the training process of the proposed DQN-based navigation system was evaluated in a goal-reaching scenario. This scenario was characterized by a fixed goal position set in an environment free of obstacles. Figure \ref{fig:reward} presents the learning progression when the goal is set at coordinates $(1, 3)$, with the vine robot starting from initial parameters $s_0 = 1$ meter and $\kappa_0 = 0.01$ m$^{-1}$.

The left side of Figure \ref{fig:reward} illustrates a notable decrease in the number of steps required for the robot to successfully complete the reaching phase. This reduction indicates an improvement in the efficiency of the robot's path-finding as the training progresses. On the right side, the figure depicts an increase in the average accumulated reward over the course of the training steps. 

It is important to note the observable fluctuations in both the number of steps and the average reward, particularly in the early stages of training. These fluctuations can be attributed to the high exploration rate $\epsilon$ initially set in the training process. At this stage, the robot is more inclined to make random decisions, exploring a broader range of policies. This exploratory approach is crucial for the robot to discover and learn from diverse situations, even though it may temporarily deviate from the policy that maximizes expected rewards. As training progresses and the exploration rate decreases, the robot increasingly adopts strategies that focus on reward maximization, leading to more consistent and goal-oriented behavior.

\begin{figure}[!t!p]
	\centering
	\includegraphics[width=.7\columnwidth,keepaspectratio]{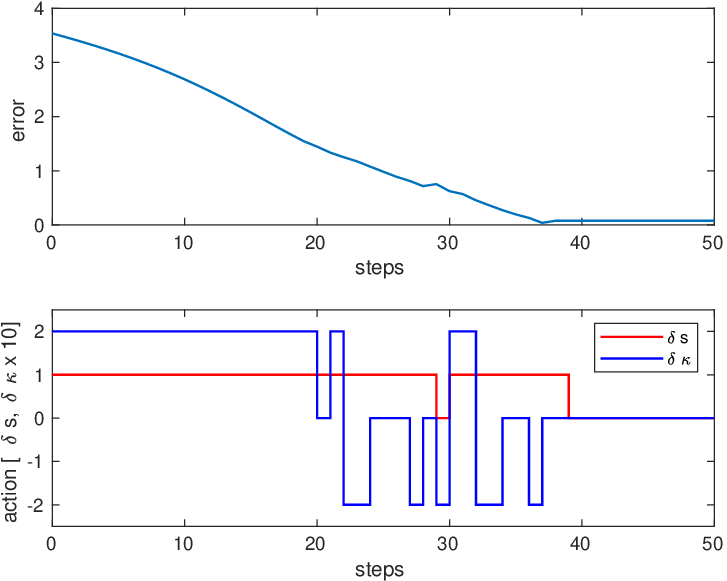}
	\caption{(Top) The decay of the error between the robot's tip and the target location of the trained DQN. (Bottom) The set of discrete action commands executed by the vine robot to reach the goal.  }
	\label{fig:decay}	
\end{figure}

Figure \ref{fig:decay} evaluates the DQN algorithm post-training, focusing on its ability to direct the robot towards a specified goal using the learned DQN model. The upper section of the figure graphically represents the decrease in error distance, which is the spatial discrepancy between the robot's tip and the designated goal. The lower section of the figure depicts the discrete control actions executed by the robot during its journey to the goal. These actions are a direct result of the decisions made by the DQN agent based on the trained model. A key observation from the results is that after 38 steps, equivalent to 3.8 seconds, the robot successfully reaches the goal, where the DQN agent applying zero control actions. 

\subsection{Varying goal, obstacle-free}
\begin{figure}[!t!p]
	\centering
	\includegraphics[width=\columnwidth,keepaspectratio]{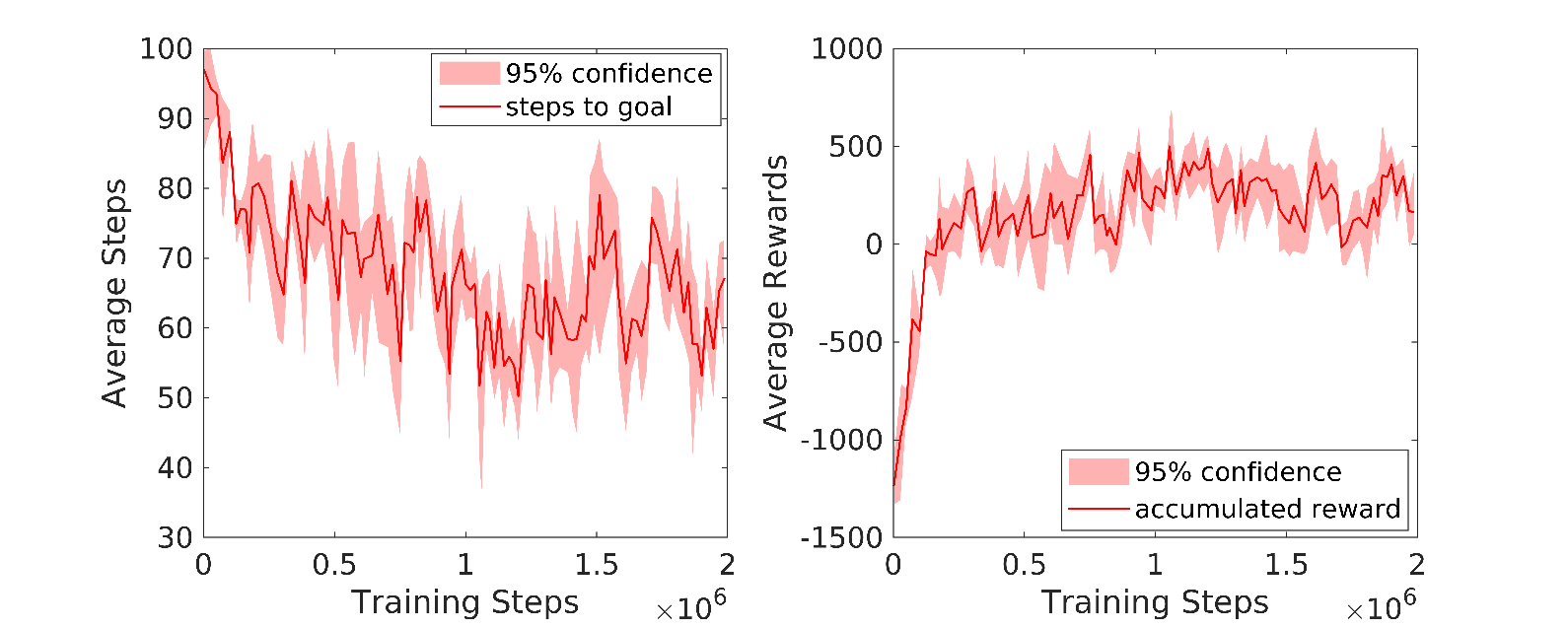}
	\caption{The learning performance of the DQN algorithm in an obstacle-free goal changing environment.}
	\label{fig:rewardGC}	
\end{figure}

To assess the generalization capabilities of the proposed DQN model, particularly in navigating towards a goal with varying location within the environment, we incorporated a goal-changing strategy during the training phase. This strategy involved altering the goal's location with a predefined probability of $10\%$, thus presenting the agent with diverse navigation challenges. The environment was kept free of obstacles for this phase of training to simplify the evaluation process and focus on the agent's ability to adapt to changing goals.

Figure \ref{fig:rewardGC} illustrates the training progress of the DQN agent under this variable goal location condition. Notably, the figure demonstrates a decreasing trend in the number of steps required for the agent to reach the goals as the episodes progress. Concurrently, there is an observable increase in the accumulated rewards over the episodes, indicating an improvement in the agent's efficiency and goal-reaching capability.

The observed fluctuations in both the number of steps and the accumulated rewards can be attributed to the dynamic nature of the goal locations. Despite these variations in goal locations, the DQN model exhibits a notable robustness over the episodes.
\begin{figure}[!b!p]
	\centering
	\includegraphics[width=\columnwidth,keepaspectratio]{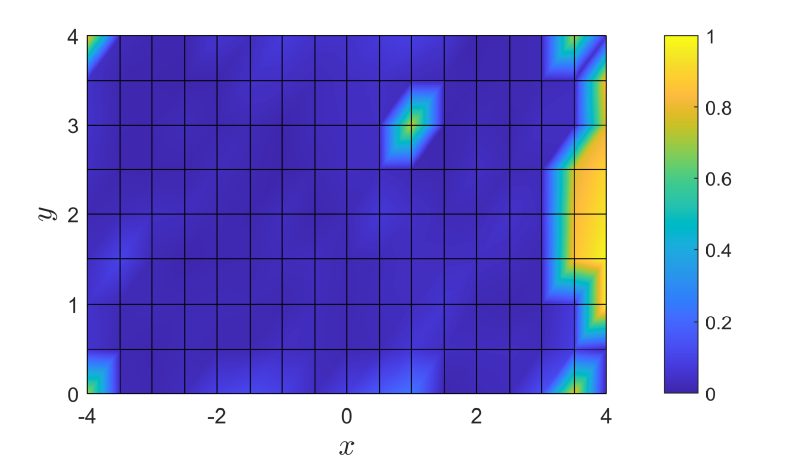}
	\caption{Normalized error distribution among the working environment of the growing robot.}
	\label{fig:error_dist}	
\end{figure}
During the testing phase, the DQN agent was tasked with directing the vine robot to reach various goals, with the goal locations $(x_o, y_o)$ sequentially altered within the environment, adhering to the ranges $x_o \in [0, 4]$ and $y_o \in [-4, 4]$. Figure \ref{fig:error_dist} presents a heat map illustrating the normalized error distribution across the working environment. The visualization indicates minimal errors in goal-reaching accuracy across diverse goal locations, with the exception of areas near the edges. This observation of increased error at the periphery could potentially be attributed to the constrained size of the simulated vine robot, which was limited to five meters.

\subsection{Varying goal, obstacle-aware}
\begin{figure}[!t!p]
	\centering
	\includegraphics[width=\columnwidth]{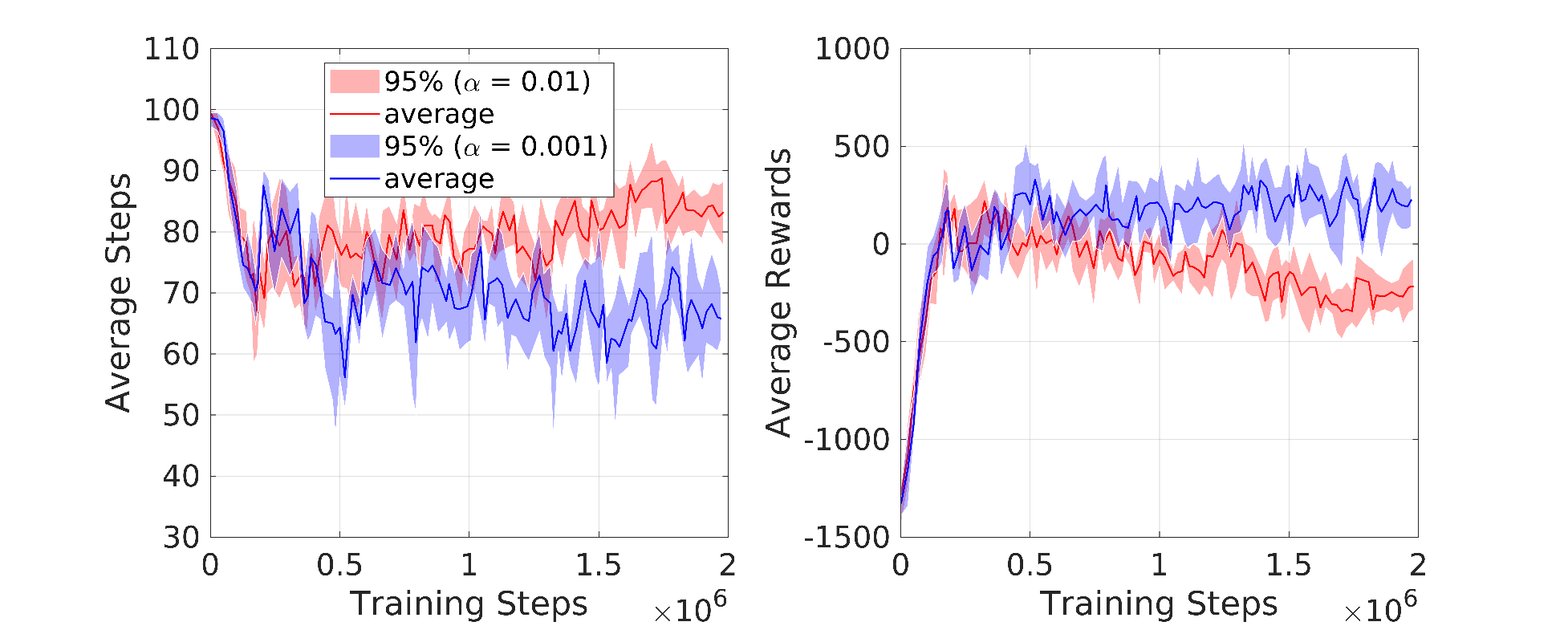}
	\caption{The training progress in terms of the number of steps and the accumulated rewards of the DQN agent for commanding the vine robot to reach a varying goal while considering the interaction with obstacles.}
	\label{fig:rewardGCalpha}	
\end{figure}

In the final series of experiments, the DQN agent underwent extensive training, consisting of 2 million steps, within an environment that included a fixed obstacle with a radius of $r_o = 0.5$ meters, located at coordinates $(1, 2)$. Concurrently, the goal's position was dynamically altered throughout the training, with a 10\% probability of being randomly relocated within the environmental boundaries. Figure \ref{fig:rewardGCalpha} illustrates the agent's progressive learning curve, showcasing an efficient reduction in the number of steps required to reach the goal while simultaneously maximizing the accumulated rewards.

This training efficacy was further analyzed under two distinct learning rate scenarios, $\alpha_1 = 0.01$ and $\alpha_2 = 0.001$, as demonstrated in Figure \ref{fig:rewardGCalpha}. The comparative analysis suggests that a lower learning rate tends to enhance the training performance, potentially leading to a more precise and refined learning outcome. However, it's important to note that this improved performance might come at the cost of increased computational resources. This trade-off between learning rate and computational demand is a crucial consideration in the optimization of the DQN training process, particularly in complex environments where obstacle navigation and dynamic goal-seeking are key factors.

\begin{figure}[!t!p]
	\centering
	\includegraphics[width=\columnwidth,keepaspectratio]{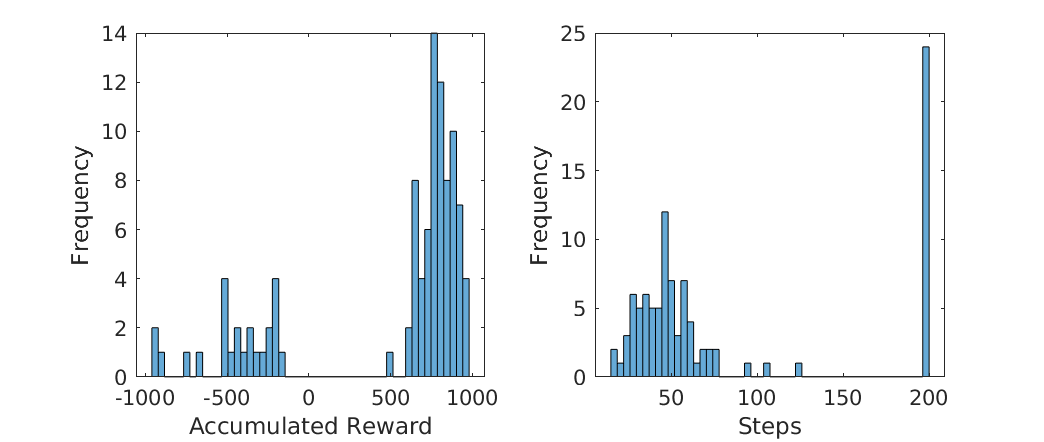}
	\caption{The histogram of the DQN agent performance in terms of the number of steps and the accumulated rewards in the varying goal obstacle-aware scenario.}
	\label{fig:hist}	
\end{figure}

During the testing phase, the proficiency of the DQN agent was rigorously evaluated through a series of 100 trials, wherein it was tasked to navigate to randomly set goals within a maximum limit of 200 time steps. The performance outcomes of these trials are comprehensively visualized in the histograms presented in Figure \ref{fig:hist}, focusing on two key metrics: the number of steps taken and the accumulated rewards.

The data illustrates that in approximately 75\% of the trials, the robot successfully accrued positive rewards ranging between 500 and 1000. This result indicates a high level of efficiency in reaching the goals. Additionally, it was observed that the majority of these successful trials were completed within 50 steps, further underscoring the agent's effectiveness.

However, it is noteworthy that in about 25\% of the trials, the robot failed to reach the goal within the stipulated 200 steps. This failure could be attributed to a couple of factors: Firstly, the random goal locations might occasionally be generated within the obstacle's space, rendering them physically inaccessible. Secondly, the goals may be positioned at distant locations in the environment, posing a challenge for the robot to reach them within the 200-step limit.

To mitigate this issue and decrease the failure rate, a couple of strategies could be implemented: One approach would be to modify the goal generation algorithm to ensure that goals are not placed within or too close to the obstacle's location. Alternatively, extending the time limit beyond 200 steps could provide the robot with additional leeway to reach more distantly placed goals. Either of these adjustments could enhance the overall success rate of the DQN agent in effectively reaching the goals across a variety of environmental conditions.

\begin{figure*}[!t!p]
	\centering
	\begin{minipage}{.33\textwidth}
		\centering
		\subfloat[$t= 0 s$]{\label{fig:traj3}\includegraphics[width=\columnwidth]{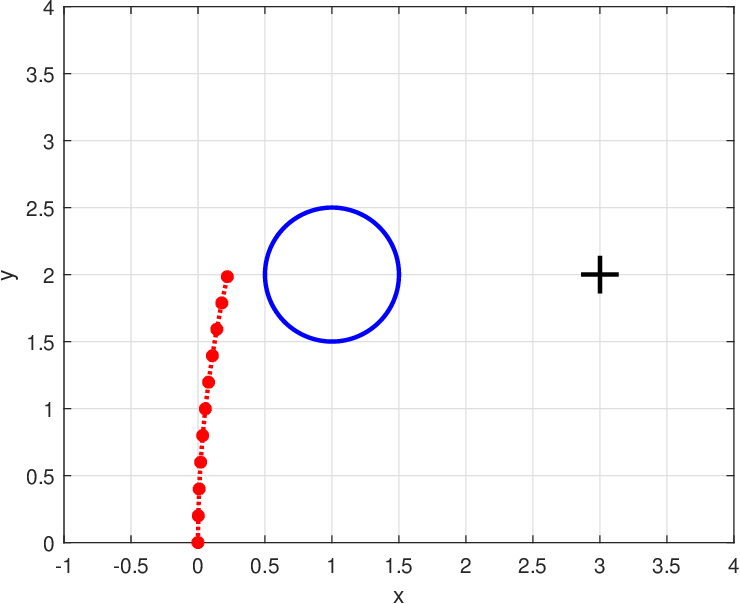}}
	\end{minipage}%
	~
	\begin{minipage}{.33\textwidth}
		\centering
		\subfloat[$t= 1.6 s$]{\label{fig:traj1}\includegraphics[width=\columnwidth]{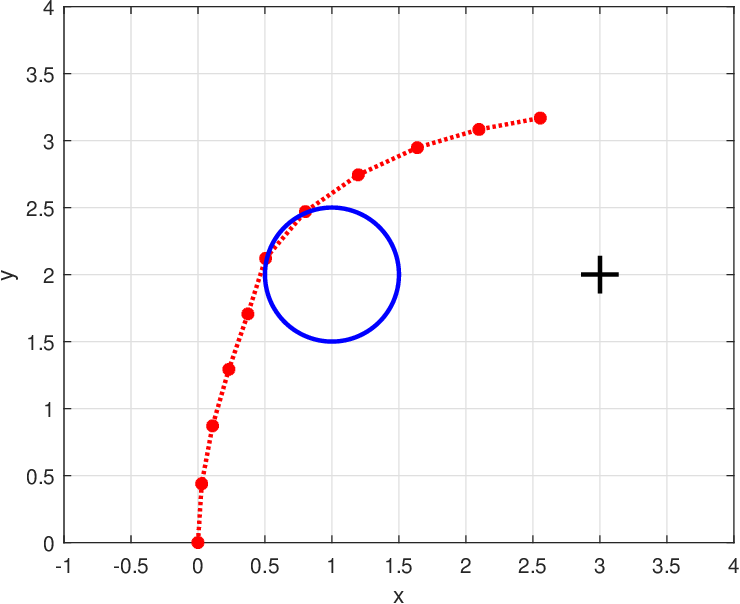}}
	\end{minipage}%
	~
	\begin{minipage}{.33\textwidth}
		\centering
		\subfloat[$t= 3.2 s$]{\label{fig:traj2}\includegraphics[width=\columnwidth]{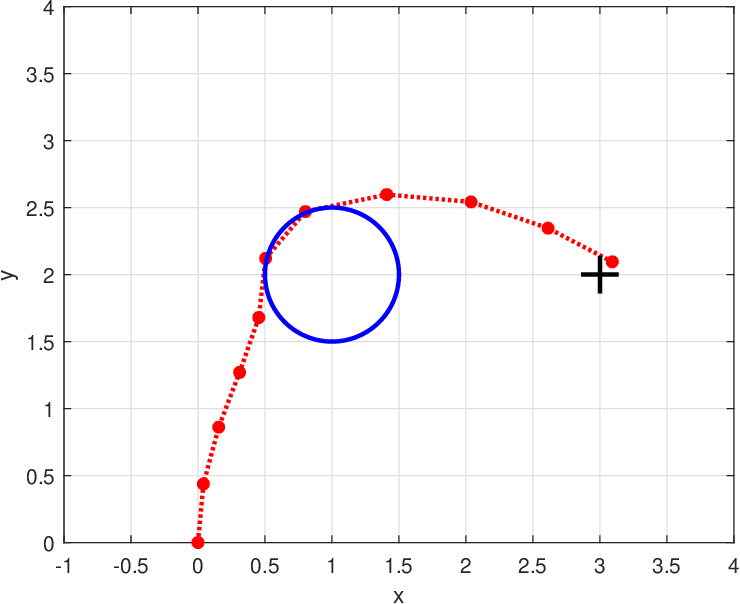}}
	\end{minipage}
\caption{The trained DQN agent successfully commands the vine growing robot to reach the goal (represented in a black cross) while considering the interaction with the obstacle located at $(1, 2)$.}
\label{fig:progress}		
\end{figure*}

Figure \ref{fig:progress} depicts a specific testing scenario for the DQN agent, illustrating a strategic navigation task involving the vine robot, marked with red indicators, and an obstacle, represented by a blue circle. This scenario demonstrates the robot's ability to exploit the presence of the obstacle in order to reach a goal that is otherwise challenging to access. The figure sequentially portrays the robot's movement across different time steps, highlighting its adaptive navigation strategy.

A key observation from this scenario is the alteration in the robot's shape upon encountering the obstacle, deviating from the constant curvature model initially assumed. This deviation is not just a consequence of the robot's interaction with the obstacle but is also strategically advantageous. The vine robot's inherent compliance, a fundamental characteristic of soft robotics, is effectively utilized here to navigate around the obstacle. This flexibility allows the robot to mold its path and shape in response to environmental constraints, showcasing an advanced level of adaptability.

The scenario illustrated in Figure \ref{fig:progress} exemplifies the practical benefits of the robot's compliance in real-world applications. It demonstrates the robot's capacity to dynamically adjust its form and trajectory in complex environments, leveraging obstacles not as hindrances, but as aids in navigating towards difficult-to-reach goals. This ability to intelligently interact with and utilize environmental features underscores the sophistication and potential of the DQN-guided navigation strategy in soft robotics.

\section{Conclusion}
\label{sec:conc}
In this research a Deep Q Network (DQN)-based navigation approach is implemented for a soft vine-growing robot resulting in significant insights into the capabilities and adaptability of this technique in environments with obstacles. Through a series of meticulously designed experiments, the study has demonstrated the proficiency of the trained DQN agent in guiding the robot to effectively reach designated goals, even in scenarios with challenging obstacle placements.

Particularly notable is the robot's ability to navigate around and utilize obstacles as part of its pathfinding strategy. This scenario illustrated the robot's sophisticated interaction with the environment, where it skillfully adapted its shape in response to obstacles, deviating from the constant curvature model. This adaptability, facilitated by the robot's inherent compliance, proved to be advantageous in navigating through and leveraging the environment to reach difficult goals.

The methodology implemented in this study adopts a discrete set of actions to accomplish the navigation task of the vine growing robot. This approach is characterized by its simplicity and ease of implementation, as it allows the robot to choose from a predefined set of actions at each step. However, this discrete action space may limit the robot's ability to perform fine-grained movements, which can be crucial in tasks requiring high precision.

As a natural progression of this research, it is essential to explore and analyze how this discrete action space approach compares with continuous space learning techniques. Continuous action spaces, as employed in algorithms like Asynchronous Advantage Actor-Critic (A3C) \cite{moore2016generalized} and Deep Deterministic Policy Gradient (DDPG) \cite{lillicrap2015continuous}, offer a broader range of potential actions at each step. This can lead to more nuanced and precise control strategies, potentially enhancing the robot's performance in complex tasks.

\bibliographystyle{IEEEtran}
\bibliography{references}

\end{document}